\documentclass[11pt,a4paper,anonymous]{article}

\usepackage[utf8]{inputenc}
\usepackage[T1, T2A]{fontenc}
\usepackage[english]{babel}
\usepackage{graphicx}
\usepackage{caption}
\usepackage{subcaption}
\usepackage{geometry}
\usepackage{sidecap}
\usepackage{array}
\usepackage{float}
\usepackage[thinlines]{easytable}
\usepackage{fancyhdr}
\usepackage{amsmath,amsfonts,amssymb,amsthm}
\usepackage{mathtools}
\usepackage{commath}
\usepackage[sc,osf]{mathpazo}
\usepackage[]{algorithm}
\usepackage{algorithmic}
\usepackage{gensymb}
\usepackage{multirow}
\usepackage{color}
\usepackage{bm}
\usepackage{breakcites}

\usepackage{nomencl}

\makenomenclature

\usepackage{etoolbox}
\renewcommand\nomgroup[1]{%
  \item[\bfseries
  \ifstrequal{#1}{A}{Symbols}{%
  \ifstrequal{#1}{B}{Greek symbols}{%
  \ifstrequal{#1}{C}{Subscripts}{}}}%
]}

\newcommand{\added}[1]{{\bf\textcolor{blue}{#1}}}

\graphicspath{ {Slike/} }

\usepackage{hyperref}
\hypersetup{
    colorlinks=true, 
    linktoc=all,     
    linkcolor=black,  
}

\title{Gaussian Conditional Random Fields for Classification}
\author{Andrija Petrović$^{a*}$\and Mladen Nikolić$^b$\and Miloš Jovanović$^a$\and Boris Delibašić$^a$}

\date{}

\begin{document}
\maketitle

$^a$\textit{ Faculty of Organizational Sciences, University of Belgrade, Center for Business Decision Making, Jove Ilica 154, 11000 Belgrade, Serbia}

$^b$ \textit{ Faculty of Mathematics, University of Belgrade, Department of Computer Science, Studentski trg 16, 11000 Belgrade, Serbia}

\section*{Abstract}
Gaussian conditional random fields (GCRF) are a well-known used structured model for continuous outputs that uses multiple unstructured predictors to form its features and at the same time exploits dependence structure among outputs, which is provided by a similarity measure. In this paper, a Gaussian conditional random fields model for structured binary classification (GCRFBC) is proposed. The model is applicable to classification problems with undirected graphs, intractable for standard classification CRFs. The model representation of GCRFBC is extended by latent variables which yield some appealing properties. Thanks to the GCRF latent structure, the model becomes tractable, efficient and open to improvements previously applied to GCRF regression models.  In addition, the model allows for reduction of noise, that might appear if structures were defined directly between discrete outputs.
Additionally, two different forms of the algorithm are presented:
GCRFBCb (GCRGBC - Bayesian) and GCRFBCnb (GCRFBC - non Bayesian). The extended method of local variational approximation of sigmoid function is used for solving empirical Bayes in Bayesian GCRFBCb variant, whereas MAP value of latent variables is the basis for learning and inference in the GCRFBCnb variant. The inference in GCRFBCb is solved by Newton-Cotes formulas for one-dimensional integration. Both models are evaluated on synthetic data. It was shown that both models achieve better prediction performance than unstructured predictors. Furthermore, computational and memory complexity is evaluated. Advantages and disadvantages of the proposed GCRFBCb and GCRFBCnb are discussed in detail.

\vspace{2mm}
\textbf{Keywords}: Structured classification, Gaussian conditional random fields, Empirical Bayes, Local variational approximation

\vfill

$^*$ \textit{Corresponding author: email: aapetrovic@mas.bg.ac.rs, tel.: +381 62 295 278}

\newpage

\section{Introduction}
\label{Sec:Introduction}

Increased quantity and variety of sources of data with correlated outputs, so called structured data, created an opportunity for exploiting
additional information between dependent outputs to achieve better prediction performance. An extensive review on topic of binary and multi-label classification with structured output is provided in \cite{hongyu}. The structured classifiers were compared in terms of accuracy and speed.

One of the most successful probabilistic models for structured output classification problems are conditional
random fields (CRF) \cite{sutton2006introduction}. CRFs were successfully applied on a variety of different structured tasks, such as: low-resource named entity recognition \cite{cotterell2017low}, image segmentation \cite{zhang2015hierarchical}, chord recognition \cite{masada2017chord} and word segmentation \cite{zia2018urdu}. The main advantages of CRFs lies in their discriminatory nature, resulting in the relaxation of independence assumptions and the label bias problem that are present in many graphical models. Additionally, availability of exact gradient evaluation and probability information made CRFs widely used in different applications. Aside of many advantages, CRFs have also many drawbacks.
Gradient computation and partition function evaluation can be computationally costly, especially for large number of feature functions. That is the reason why CRFs can be very computationally expensive during inference and learning, and consequently slow.
Moreover, the CRFs with complex structure usually do not support decode-based learning \cite{sunconditional}. Sometimes even the gradient computation is impossible or exact inference is intractable due to complicated partition function.

In order to solve these problems, a wide range of different algorithms have been developed and adapted for various task. The mixtures of CRFs capable to model data  that come from multiple different sources or domains is presented in \cite{kim2017mixtures}.  The method is related to the well known hidden-unit CRF (HUCRF) \cite{maaten2011hidden}.  The conditional likelihood and expectation minimization (EM) procedure for learning have been derived there. The mixtures of CRF models were implemented on several real-world applications resulting in prediction improvement. Recently, the model based on unification of deep learning and CRF was developed by \cite{chen2016word}. The deep CRF model showed better performance compared to either shallow CRFs or deep learning methods on their own. Similarly, the combination of CRFs and deep convolutional neural networks was evaluated on an example of environmental microorganisms labeling \cite{kosov2018environmental}. The spatial relations among outputs were taken in consideration and experimental results have shown satisfactory results.

Structured models for regression based on CRFs have recently been a focus of many researchers. One of the popular methods for structured regression -- Gausian conditional random fields (GCRF) -- has the form of multivariate Gaussian distribution. The main assumption of the model is that the relations between outputs are presented in quadratic form. The multivariate Gaussian distribution representation of a CRF has many advantages, like convex loss function and, consequently, efficient inference and learning.

The GCRF model was first implemented for the task of low-level computer vision \cite{tappen2007learning}. Since than, various different adaptations and approximations of GCRF were proposed \cite{radosavljevic2014neural}. The parameter space for the GCRF model is extended to facilitate joint modelling of positive and negative influences \cite{glass2016extending}. In addition, the model is extended by bias term into link weight and solved as a part of convex optimization. Semi-supervised model marginalized Gaussian conditional random fields (MGCRF) for dealing with missing variables were proposed by \cite{stojanovic2015semi}. The benefits of the model were proved on partially observed data and showed better prediction performance then alternative semi-supervised structured models.

In this paper, a new model of Gaussian conditional random fields for binary classification is proposed (GCRFBC). The model assumes that discrete outputs $y_i$ are conditionally independent for given continuous latent variables $z_i$ which follow a distribution modeled by a GCRF. That way, relations between discrete outputs are not expressed directly. Two different inference and learning approaches are proposed in this paper. The first one is based on evaluating empirical Bayes by marginalizing latent variables (GCRFBCb), whereas MAP value of latent variables is the basis for learning and inference in the second model (GCRFBCnb). The presented models are tested on synthetic data. This is a discrete output problem, so it is not possible to use standard GCRFs for regression.

In section~\ref{Sec:Related work} the related work is reviewed and the GCRF model is briefly presented. The details of the proposed models along with the inference and learning are described in section~\ref{Sec:Methodology}. Experimental results on synthetic data and real-world applications are shown in section~\ref{Sec:Experiments}. Final conclusions are given in section~\ref{Sec:Conclusion}.

\section{Related Work and Background Material}
\label{Sec:Related work}
GCRF is a discriminative graph-based regression model \cite{radosavljevic2010continuous} . Nodes of the graph are variables $\bm{y} = \left(y_1, y_2, \ldots, y_N \right)$, which need to be predicted  given a set of features $\bm{x}$. The attributes $\bm{x}=\left(\bm{x_1}, \bm{x_2}, \ldots, \bm{x_N}\right)$ interact with each node $y_i$ independently of one another, while the relations between outputs are expressed by pairwise interaction function. In order to learn parameters of the model, a training set of vectors of attributes $x$ and real-valued response variables $y$ are provided. The generalized form of the conditional distribution $P\left(\bm{y}|\bm{x},\bm{\alpha},\bm{\beta} \right)$ is:
\begin{equation}
P\left(\bm{y}|\bm{x},\bm{\alpha},\bm{\beta}\right)=\frac{1}{Z\left(\bm{x},\bm{\alpha},\bm{\beta}\right)}\textrm{exp}\left(\sum_{i=1}^{N} A(\bm{\alpha}, y_i, \bm{x_i})+\sum_{i \neq j} I(\bm{\beta}, y_i, y_j)\right)
\end{equation}

Two different feature functions are used: \textit{association potential} $A(\alpha,y_i,\bm{x})$ to model relations between outputs $y_i$ and corresponding input vector $\bm{x_i}$ and \textit{interaction potential} $I(\bm{\beta}, y_i, y_j)$ to model pairwise relations between nodes. Vectors $\bm{\alpha}$ and $\bm{\beta}$ are parameters of the association potential $A$  and the interaction potential $I$, whereas $Z$ is partition function. The association potential is defined as:
\begin{equation}
A(\bm{\alpha}, y_i, \bm{x_i}) = -\sum_{k=1}^{K}\alpha_k\left(y_i-R_k \left(\bm{x_i}\right)\right)^2
\end{equation}
where $R_k (\bm{x_i})$ represents unstructured predictor of $y_i$ for each node in the graph. This unstructured predictor can be any regression model that gives prediction of output $y_i$ for given attributes $\bm{x_i}$. $K$ is the total number of unstructured predictors. The interaction potential functions are defined as:
\begin{equation}
I(\bm{\beta}, y_i, y_j) = -\sum_{l=1}^{L}\sum_{k=1}^{K}\beta_lS_{ij}^l(y_i-y_j)^2
\end{equation}
where $S_{ij}^l$ is value that express similarity between nodes $i$ and $j$ in graph $l$. $L$ is the total numbers of graphs (similarity functions). Graphs can express any kind of relations between nodes e.g., spatial and temporal correlations between outputs. One of the main advantages of GCRF is the ability to express different relations between outputs by variety of graphs. Moreover, the GCRF is able to learn which graph is significant for outputs prediction.


The quadratic form of interaction and association potential enables conditional distribution $P(\bm{y}|\bm{x},\bm{\alpha},\bm{\beta})$ to be expressed as multivariate Gaussian distribution. The canonical form of GCRF is \cite{radosavljevic2010continuous}:
\begin{equation}
\label{GCRF2}
P(\bm{y}|\bm{x},\bm{\alpha},\bm{\beta})=\frac{1}{(2\pi)^{\frac{N}{2}}|\Sigma|^{\frac{1}{2}}}\textrm{exp}\left(-\frac{1}{2}(\bm{y}-\bm{\mu})^T\Sigma^{-1}(\bm{y}-\bm{\mu})\right)
\end{equation}
where $\left|\cdot\right|$ denotes determinant. Precision matrix $\Sigma^{-1}=2Q$ and distribution mean $\bm{\mu}=\Sigma\bm{b}$ is defined as, respectively:
\begin{equation}
 Q= \begin{cases}
\sum_{k=1}^{K} \alpha_k+\sum_{h=1}^{N}\sum_{l=1}^{L}\beta_lS_{ih}^l, & \text{if $i=j$} \\
-\sum_{l=1}^{L}\beta_lS_{ij}^l, & \text{if $i \neq j$}
  \end{cases}
\end{equation}
\begin{equation}
b_i=2\left(\sum_{k=1}^K\alpha_kR_k(\bm{x_i})\right)
\end{equation}

Due to concavity of multivariate Gaussian distribution the inference task $\underset{\bm{y}}{\text{argmax}}P(\bm{y}|\bm{x},\bm{\alpha},\bm{\beta})$ is straightforward. The maximum posterior estimate of $\bm{y}$ is the distribution expectation $\bm{\mu}$. \\
The objective of the learning task is to optimize parameters $\bm{\alpha}$ and $\bm{\beta}$ by maximizing conditional log likelihood  $\underset{\bm{\alpha}, \bm{\beta}}{\text{argmax}}\sum_{\bm{y}}\textrm{log}P(\bm{y}|\bm{x},\bm{\alpha},\bm{\beta})$. One way to ensure positive definiteness of the covariance matrix of GCRF  is to require diagonal dominance \cite{strang1993introduction}. This can be ensured by imposing constraints that all elements of $\bm{\alpha}$ and $\bm{\beta}$ be greater than 0 \cite{radosavljevic2010continuous}.

Large number of different studies connected with graph based methods for regression can be found in the literature \cite{fox2015applied}. A comprehensive review of continuous conditional random fields (CCRF) was provided in \cite{radosavljevic2010continuous}. The sparse conditional random fields obtained by $l_1$ regularization are first proposed and evaluated by \cite{wytock2013sparse}. Additionaly, \cite{frot2018graphical} presented GCRF with the latent variable decomposition and derived convergence bounds for the estimator that is well behaved in high dimensional regime.

One of the adapatations of GCRF on discrete output was briefly discussed in \cite{radosavljevic2011gaussian}, as a part of future work directions that should be considered. Namely, the model should assume existence of latent continuous variables that follows distribution modeled by GCRF, whereas the distribution of discrete outputs $y$ is in multivariate normal distribution form with diagonal covariance matrix. The parameters of the models are obtained by EM algorithm. Moreover, since $\bm{y}$ and $\bm{z}$ are unknown, the inference is performed by first calculating marginal expectation for $\bm{z}$. Discrete values of outputs are found as average values of $\bm{z}$ over positive and negative examples. The models developed in this paper are motivated by the preceding discussion.

\section{Methodology}
\label{Sec:Methodology}
One way of adapting GCRF to classification problem is by approximating discrete outputs by suitably defining continuous outputs. Namely, GCRF can provide dependence structure over continuous variables which can be passed through sigmoid function. That way relationship between regression GCRF and classification GCRF is similar to the relationship between linear and logistic regression, but with dependent variables. Aside from allowing us to define a classification variant of GCRF, this may result in  additional appealing properties:

\begin{itemize}
\item The model is applicable to classification problems with undirected graphs, intractable for standard classification CRFs. Thanks to the GCRF latent structure, the model becomes tractable, efficient and open to improvements previously applied to GCRF regression models.
\item Defining correlations directly between discrete outputs may introduce unnecessary noise to the model \cite{tan2010social}. This problem can be solved by defining structured relations on a latent continuous variable space.
\item In case that unstructured predictors are unreliable, which is signaled by their large variance (diagonal elements in the covariance matrix), it is simple to marginalize over latent variable space and obtain better results.
\end{itemize}

\begin{figure}[H]
    \centering
        \includegraphics[angle=0, width=0.6\textwidth]{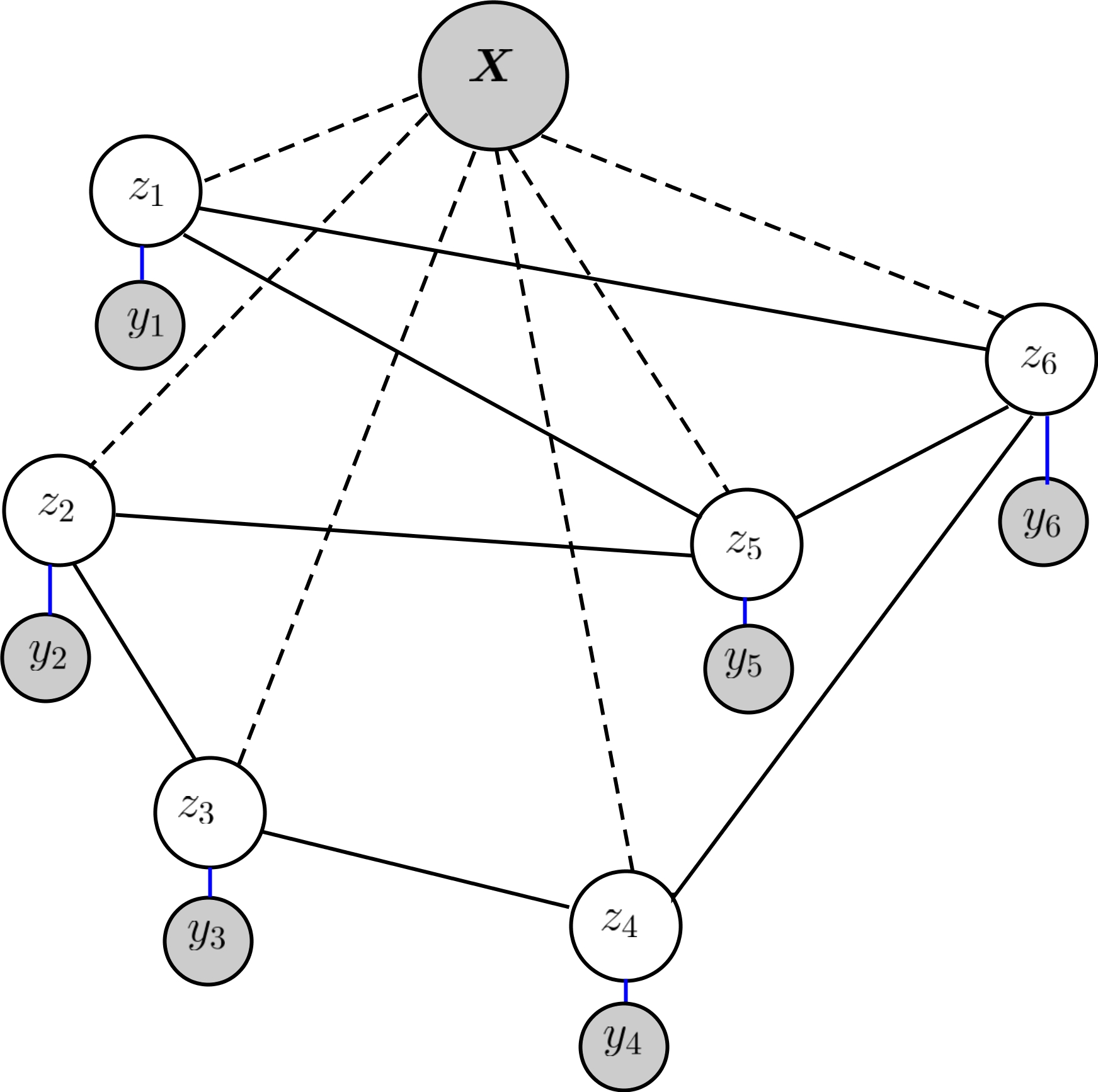}
        \caption{Graphical representation of dependencies expressed by GCRFBC model}
        \label{fig:Fig1}
\end{figure}

It is assumed that $y_i$ are discrete binary outputs and $z_i$ are continuous latent variables assigned to each $y_i$. In addition, each output $y_i$ is conditionally independent of the others given $z_i$. The illustration of dependencies expressed by GCRFBC model is presented in Fig.~\ref{fig:Fig1}.

The conditional probability distribution $P(y_i|z_i)$ is defined as Bernoulli distribution:
\begin{equation}
P(y_i|z_i)=Ber(y_i|\sigma(z_i))=\sigma(z_i)^{y_i}(1-\sigma(z_i))^{1-y_i}
\end{equation}
where $\sigma(\cdot)$ is sigmoid function. Due to conditional independence assumption, the joint distribution of outputs $y_i$ can be expressed as:
\begin{equation}
P(y_1,y_2,\ldots , y_N|\bm{z})=\prod_{i=1}^N\sigma(z_i)^{y_i}(1-\sigma(z_i))^{1-y_i}
\end{equation}
Furthermore, the conditional distribution $P(\bm{z}|\bm{x})$ is the same as in the classical GCRF model and has canonical form defined by multivariate Gaussian distribution. Hence, joint distribution of continuous latent variables $\bm{z}$ and outputs $\bm{y}$ is:

\begin{equation}
\label{Reprezentacija}
\begin{split}
P(\bm{y,z}|\bm{x},\bm{\theta})=\prod_{i=1}^N\sigma(z_i)^{y_i}(1-\sigma(z_i))^{1-y_i}& \cdot \frac{1}{(2\pi)^{N/2}\left|\Sigma(\bm{x})\right| ^{1/2}} \\
&\cdot
\textrm{exp}\left(-\frac{1}{2}(\bm{z}-\bm{\mu(\bm{x})})^T\Sigma(\bm{x}) ^{-1}(\bm{z}-\bm{\mu(\bm{x})})\right)
\end{split}
\end{equation}
where $\bm{\theta}=(\alpha_1,\ldots,\alpha_K, \beta_1,\ldots,\beta_L)$.

Equation~\ref{Reprezentacija} presents the general form of mathematical model representation that will be further discussed in this paper.

We consider two ways of inference and learning in GCRFBC model:
\begin{itemize}
\item GCRFBCb - with conditional probability distribution $P(\bm{y}|\bm{x},\bm{\theta})$, in which variables $\bm{z}$ are marginalized over, and
\item GCRFBCnb - with conditional probability distribution $P\left(\bm{y}|\bm{x},\bm{\theta},\mu_{\bm{z}}\right)$, in which variables $\bm{z}$ are substituted by their expectations.
\end{itemize}

\subsection{Inference}
\label{Sec:Inference}

\subsubsection{Inference in GCRFBCb Model}
Prediction of discrete outputs $\bm{y}$ for given features $\bm{x}$ and parameters $\bm{\theta}$ is analytically intractable due to integration of the joint distribution $P(\bm{y,z}|\bm{x},\bm{\theta})$ with respect to latent variables. However, due to conditional independence between nodes, it is possible to obtain $P(y_i=1|\bm{x},\bm{\theta})$.

\begin{equation}
P(y_i|\bm{x},\bm{\theta})=\int_{\bm{z}} P(y_i|\bm{z})P(\bm{z}|\bm{x},\bm{\theta}) d\bm{z}\\
\end{equation}

\begin{equation}
P(y_i=1|\bm{x},\bm{\theta})=\int_{\bm{z}} \sigma(z_i)P(\bm{z}|\bm{x},\bm{\theta}) d\bm{z}
\end{equation}

\sloppy As a result of independence properties of the distribution, it holds $P(y_i=1|\bm{z})=P(y_i=1|z_i)$, and it is possible to marginalize $P(\bm{z}|\bm{x},\bm{\theta})$ with respect to latent variables $\bm{z}\added{'}=\left(z_1,\ldots, z_{i-1}, z_{i+1},\ldots,z_N\right)$:
\begin{equation}
P(y_i=1|\bm{x},\bm{\theta})=\int_{z_i} \sigma(z_i) \left(\int_{\bm{z}'} P(\bm{z}',z_i|\bm{x},\bm{\theta})d\bm{z}'\right) d{z_i}
\end{equation}
where $ \int_{\bm{z}'} P(\bm{z}',z_i|\bm{x},\bm{\theta}) d\bm{z}'$ is normal distribution with mean $\mu=\mu_{i}$ and variance $\sigma^2_{i}=\Sigma_{ii}$. Therefore, it holds:
\begin{equation}
P(y_i=1|\bm{x},\bm{\theta})=\int_{-\infty}^{+\infty}\sigma(z_i) {\cal N}(z_i|\mu_i,\sigma^2_i)dz_i
\end{equation}
The evaluation of $P(y_i=0|\bm{x},\bm{\theta})$ is straightforward and is expressed as:
\begin{equation}
P(y_i=0|\bm{x},\bm{\theta})=1-P(y_i=1|\bm{x},\bm{\theta})
\end{equation}

\begin{figure}[H]
    \centering
        \includegraphics[angle=0, width=1\textwidth]{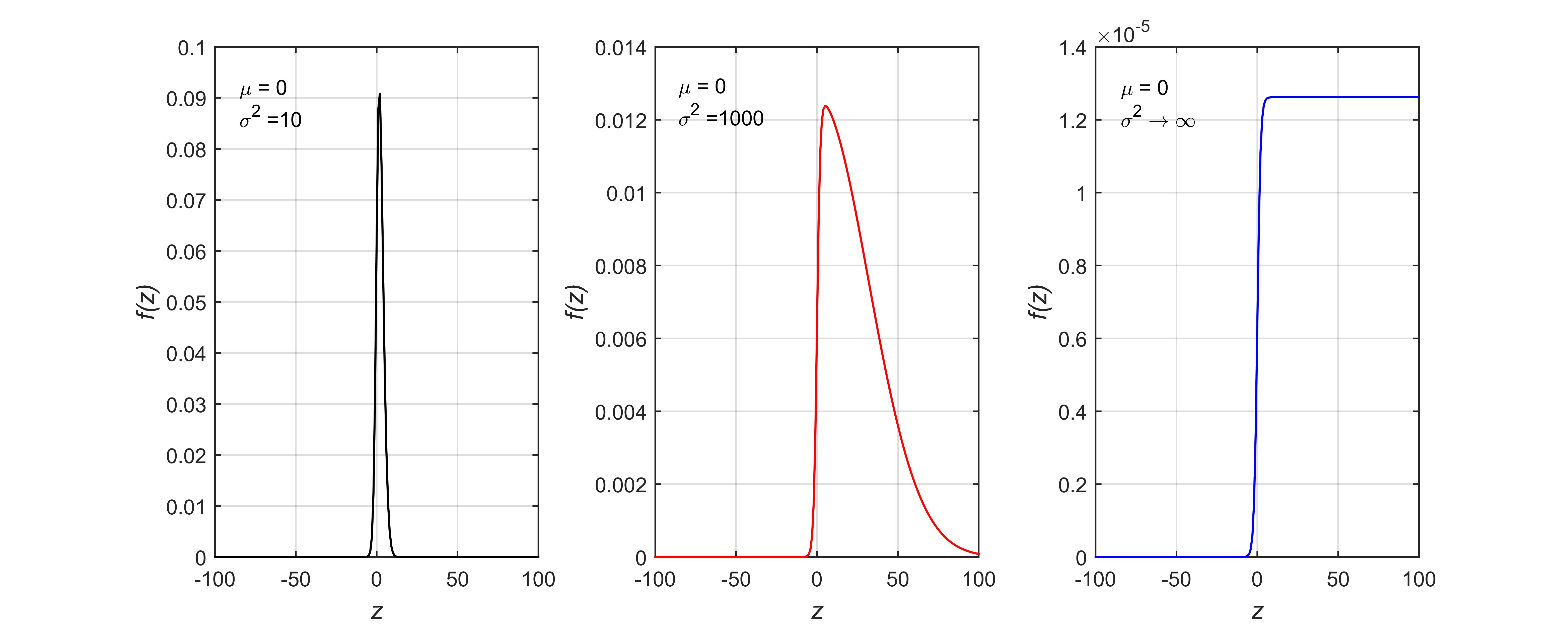}
        \captionsetup{justification=centering}
        \caption{Shapes of function $\sigma(z) N(z_i|\mu_{ii},\sigma^2)$ for three different choices of variance of the normal distribution.}
        \label{fig:Fig2}
\end{figure}

The one-dimensional integral is still analytically intractable, but can be effectively evaluated by one-dimensional numerical integration. Additionally, the surface of the function expressed by the product of the univariate normal distribution and sigmoid function is mostly concentrated closely around the mean, except in cases in which variance of normal distribution is high. The plot of function $\sigma(z) N(z_i|\mu_{ii},\sigma^2)$ with respect to the variance of normal distribution is illustrated by Fig.~\ref{fig:Fig2}. Therefore, the limits of the integral $(-\infty,+\infty)$ can be reasonably approximated by the interval $(\mu-10 \sigma_i, \mu+10 \sigma_i)$. This approximation improves integration precision, especially in case that Newton-Cotes formulas are used for numerical integration \cite{davis2007methods}. The proposed inference approach can be effectively used in case of huge number of nodes, due to low computational cost of one-dimensional numerical integration.

\subsubsection{Inference in GCRFBCnb Model}
The inference procedure in GCRFBCnb is much simpler, because marginalization with respect to latent variables is not performed. To predict $\bm{y}$, it is necessary to evaluate posterior maximum of latent variable $\bm{z}_{\max}=\underset{\bm{z}}{\text{argmax}}P(\bm{z}|\bm{x},\bm{\theta})$, which is straightforward due to normal form of GCRF. Therefore, it holds $\bm{z}_{\max}=\bm{\mu_{z,i}}$. The conditional distribution $P(y_i=1|\bm{x},\bm{\mu_{z,i}},\bm{\theta})$ can be expressed as:
\begin{equation}
P(y_i=1|\bm{x},\bm{\mu_z},\bm{\theta})=\sigma(\mu_{z,i})=\frac{1}{1+\textrm{exp}(-\mu_{z,i})}
\end{equation}
where $\mu_{z,i}$ is expectation of latent variable $z_i$.

\subsection{Learning}

\subsubsection{Learning in GCRFBCb Model}
In comparison with inference, learning procedure is  more complicated. Evaluation of the conditional log likelihood is intractable, since latent variables cannot be analytically marginalized. The conditional log likelihood is expressed as:
\begin{equation}
\begin{split}
{\cal L}\left(\bm{Y}|\bm{X},\bm{\theta}\right)=\log\left(\int_{\bm{Z}} P(\bm{Y,Z}|\bm{\theta},\bm{X})d\bm{Z}\right)&=\sum_{j=1}^M\log\left(\int_{\bm{z_j}} P(\bm{y_j,z_j}|\bm{\theta},\bm{x})d\bm{z_j}\right)\\
&=\sum_{j=1}^M{{\cal L}}_j(\bm{y_j}|\bm{x},\bm{\theta})
\end{split}
\end{equation}
\begin{equation}
\label{logLike}
{\cal L}_j(\bm{y_j}|\bm{x},\bm{\theta})=\log \int_{\bm{z_j}} \prod_{i=1}^N\sigma(z_{ji})^{y_{ji}}(1-\sigma(z_{ji}))^{1-y_{ji}}\frac{\exp(-\frac{1}{2}(\bm{z_j}-\bm{\mu_j})^T\Sigma_j ^{-1}(\bm{z_j}-\bm{\mu_j}))}{(2\pi)^{N/2}\left|\Sigma_j\right| ^{1/2}}d\bm{z_j}
\end{equation}
where $\bm{Y} \in \mathbb{R} ^{M\times N}$ is complete dataset of outputs, $\bm{X} \in \mathbb{R} ^{M\times N \times A}$ is complete dataset of features, $M$ is the total number of instances and $A$ is the total number of features. Please note that each instance is structured, so while different instances are independent of each other, variables within one instance are dependent.

\begin{figure}[H]
    \centering
        \includegraphics[angle=0, width=1\textwidth]{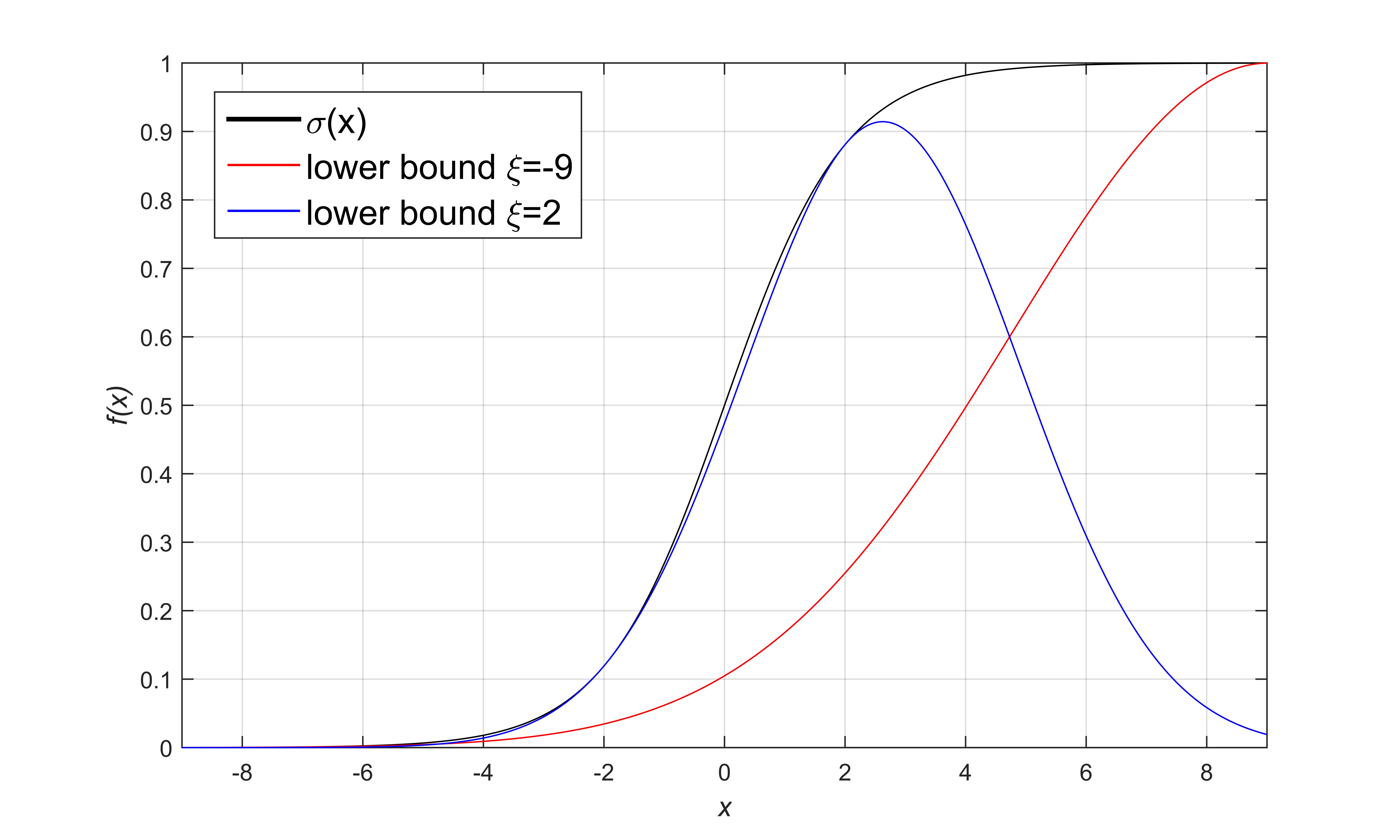}
        \caption{The sigmoid function with its lower bound}
        \label{fig:Fig3}
\end{figure}

One way to approximate integral in conditional log likelihood is by local variational approximation. \cite{jaakkola2000bayesian} derived lower bound for sigmoid function, which can be expressed as:
\begin{equation}
\label{Lbound}
\sigma(x) \geqslant \sigma(\xi) \exp \{(x-\xi)/2 - \lambda(\xi)(x^2-\xi^2)\}
\end{equation}
where $\lambda(\xi)= -\frac{1}{2 \xi} \cdot \left[ \sigma(\xi)-\frac{1}{2} \right]$ and $\xi$ is a variational parameter.
The Eq.~\ref{Lbound} is called $\xi$ \textit{transformation} of sigmoid function and it yields maximum value when $\xi=x$. The sigmoid function with lower bound is shown in Fig.~\ref{fig:Fig3}.
This approximation can be applied to the model defined by Eq.~\ref{logLike}, but the variational approximation has to be further extended because of the product of sigmoid functions, such that:
\begin{equation}
P(\bm{y_j},\bm{z_j}|\bm{\theta},\bm{x})=P(\bm{y_j}|\bm{z_j})P(\bm{z_j}|\bm{x},\bm{\theta})\geq \underline{P}(\bm{y_j},\bm{z_j}|\bm{\theta},\bm{x},\bm{\xi_j})
\end{equation}
\begin{equation}
\label{Eqprva}
\begin{split}
\underline{P}(\bm{y_j},\bm{z_j}|\bm{\theta},\bm{x}, \bm{\xi_j})=
\prod_{i=1}^N\sigma(\xi_{ji})\exp\left(z_{ji}y_{ji}-\frac{z_{ji}+\xi_{ji}}{2}-\lambda(\xi_{ji})(z^2_{ji}-\xi^2_{ji})\right)
\cdot \\
\frac{1}{(2\pi)^{N/2}\left|\Sigma_j\right| ^{1/2}}\exp\left(-\frac{1}{2}(\bm{z_j}-\bm{\mu_j})^T\Sigma_j^{-1}(\bm{z_j}-\bm{\mu_j})\right)
\end{split}
\end{equation}

The Eq.~\ref{Eqprva} can be arranged in the form suitable for integration. The lower bound of conditional log likelihood $\underline{\cal L}(\bm{y_j}|\bm{\theta},\bm{x},\bm{\xi_j})$ is defined as:
\begin{equation}
\begin{split}
\underline{{\cal L}_j}(\bm{y_j}|\bm{x_j},\bm{\theta},\bm{\xi_j})=\log\underline{P}(\bm{y_j}|\bm{x_j},\bm{\theta},\bm{\xi_j})=
&\sum_{i=1}^N\left(\log\sigma(\xi_{ji})-\frac{\xi_{ji}}{2}+\lambda(\xi_{ji})\xi^2_{ji}\right)-\\
&\frac{1}{2}\bm{\mu_j}^T\Sigma_j^{-1}\bm{\mu_j}+\frac{1}{2}\bm{m_j}^TS_j^{-1}\bm{m_j}+\frac{1}{2}\log|S_j|
\end{split}
\end{equation}
where:
\begin{equation}
S_j^{-1}=\Sigma_j^{-1}+2\Lambda_j
\end{equation}

\begin{equation}
\bm{m_j} = \Sigma_j \left( (\bm{y_j}-\frac{1}{2}\bm{I})+\Sigma_j^{-1}\bm{\mu_j} \right)
\end{equation}
\begin{equation}
\Lambda_j =
\begin{bmatrix}
    \lambda(\xi_{j1}) & 0 & 0 & \dots  & 0 \\
    0 &  \lambda(\xi_{j2}) & 0 & \dots  & 0 \\
    \vdots & \vdots & \vdots & \ddots & \vdots \\
    0 & 0 & 0 & \dots  &  \lambda(\xi_{jN})
\end{bmatrix}
\end{equation}

GCRFBCb uses the derivative of conditional log likelihood in order to find the optimal values for parameters $\bm{\alpha}$, $\bm{\beta}$ and matrix of variational parameters $\bm{\xi} \in \mathbb{R} ^{M\times N}$ by gradient ascent method. In order to ensure positive definiteness of normal distribution involved, it is sufficient to constrain parameteres $\bm{\alpha} > 0$ and $\bm{\beta} > 0$. The partial derivative of conditional log likelihood $\frac{\partial \underline{{\cal L}}_j(\bm{y_j}|\bm{x},\bm{\theta},\bm{\xi_j})}{\partial \alpha_k}$ is computed as:
\begin{equation}
\begin{split}
\frac{\partial \underline{{\cal L}}_j(\bm{y_j}|\bm{x},\bm{\theta},\bm{\xi_j})}{\partial \alpha_k} =& -\frac{1}{2}\textrm{Tr}\left(S_j \frac{\partial S_j^{-1}}{\partial \alpha_k}\right) + \frac{\partial \bm{m_j}^T}{\partial \alpha_k}S_j^{-1}\bm{m_j}+\frac{1}{2}\bm{m_j}^T\frac{\partial S_j^{-1}}{\partial \alpha_k}\bm{m_j}\\
&-\frac{\bm{\mu_j}^T}{\partial \alpha_k}\Sigma_j^{-1}\bm{\mu_j}-\frac{1}{2}\bm{\mu_j}^T\frac{\partial \Sigma_j^{-1}}{\partial \alpha_k}+\frac{1}{2}\textrm{Tr}\left(\Sigma_j\frac{\partial \Sigma_j^{-1}}{\partial \alpha_k}\right)
\end{split}
\end{equation}
where:
\begin{equation}
\label{Sigmaalfa}
\frac{\partial S_j^{-1}}{\partial \alpha_k} = \frac{\partial \Sigma_j^{-1}}{\partial \alpha_k}= \begin{cases}
2, \text{ if $i= j$}\\
0, \text{ if $i \neq j$}
\end{cases}
\end{equation}
\begin{equation}
\frac{\partial \bm{m_j^T}}{\partial \alpha_k}= -\left(\bm{y_j}-\frac{1}{2}\bm{I}+\bm{\mu_j}^T\Sigma_j^{-1}\right)S_j\frac{\partial S_j^{-1}}{\partial \alpha_k}S_j+\frac{\partial \bm{\mu_j}^T}{\partial \alpha_k}\Sigma_j^{-1}S_j + \bm{\mu_j}^T\frac{\partial \Sigma_j^{-1}}{\alpha_k}S_j
\end{equation}
\begin{equation}
\label{alpha}
\frac{\partial \mu_j^T}{\partial \alpha_k}=\left(2 \alpha_k R_k(\bm{x})- \frac{ \partial \Sigma_j^{-1}}{\partial \alpha_k}\bm{\mu_j}\right)^T\Sigma_j^T
\end{equation}
Similarly partial derivatives with respect to $\bm{\beta}$ can be defined as:
\begin{equation}
\begin{split}
\frac{\partial \underline{{\cal L}}_j(\bm{y_j}|\bm{x},\bm{\theta},\bm{\xi_j})}{\partial \beta_l} =& -\frac{1}{2}\textrm{Tr}\left(S_j \frac{\partial S_j^{-1}}{\partial \beta_l}\right) + \frac{\partial \bm{m_j}^T}{\partial \beta_l}S_j^{-1}\bm{m_j}+\frac{1}{2}\bm{m_j}^T\frac{\partial S_j^{-1}}{\partial \beta_l}\bm{m_j}\\
&-\frac{\bm{\mu_j}^T}{\partial \beta_l}\Sigma_j^{-1}\bm{\mu_j}-\frac{1}{2}\bm{\mu_j}^T\frac{\partial \Sigma_j^{-1}}{\partial \beta_l}+\frac{1}{2}\textrm{Tr}\left(\Sigma_j\frac{\partial \Sigma_j^{-1}}{\partial \beta_l}\right)
\end{split}
\end{equation}
where:
\begin{equation}
\label{Sigmabeta}
\frac{\partial S_j^{-1}}{\partial \beta_l} = \frac{\partial \Sigma_j^{-1}}{\partial \beta_l}= \begin{cases}
\sum_{n=1}^N e_{in}^lS_{in}^{l}(x), \text{ if $i= j$}\\
-e_{ij}^lS_{ij}^{l}(x), \text{ if $i \neq j$}
\end{cases}
\end{equation}
\begin{equation}
\frac{\partial \bm{m_j^T}}{\partial \beta_l}= -\left(\bm{y_j}-\frac{1}{2}\bm{I}+\bm{\mu_j}^T\Sigma_j^{-1}\right)S_j\frac{\partial S_j^{-1}}{\partial \beta_l}S_j+\frac{\partial \bm{\mu_j}^T}{\partial \beta_l}\Sigma_j^{-1}S_j + \bm{\mu_j}^T\frac{\partial \Sigma_j^{-1}}{\beta_l}S_j
\end{equation}
\begin{equation}
\label{beta}
\frac{\partial \bm{\mu_j}^T}{\partial \beta_l}=\left(- \frac{ \partial \Sigma_j^{-1}}{\partial \beta_l}\bm{\mu_j}\right)^T\Sigma_j^T
\end{equation}
In the same manner partial derivatives of conditional log likelihood with respect to $\xi_{ji}$ are:
\begin{equation}
\begin{split}
\frac{\partial \underline{{\cal L}}_j(\bm{y_j}|\bm{x},\bm{\theta},\bm{\xi_j})}{\partial \xi_{ji}}&=-\frac{1}{2}\textrm{Tr}\left(2S_j\frac{\partial \Lambda_j}{\partial \xi_{ji}}\right)-\left[2\left(\bm{y_j}-\frac{1}{2}\bm{I}\right)S_j\frac{\partial \Lambda_j}{\partial \xi_{ji}}S_j\right]S_j^{-1}\bm{m_j}\\
&+\bm{m_j}^T\frac{\partial \Lambda_j}{\partial \xi_{ji}}m_j+\sum_{i=1}^N\left(\left(\frac{1}{\sigma(\xi_{ji})}+\frac{1}{2}\xi_{ji}\right)\frac{\partial \sigma(\xi_{ji})}{\partial \xi_{ji}}+\frac{1}{2}\left(\sigma({\xi_{ji}})-\frac{3}{4}\right)\right)
\end{split}
\end{equation}
where:
\begin{equation}
\frac{\partial \Lambda_j}{\partial \xi_{ji}} =
\begin{bmatrix}
    0 & 0 & 0 & \dots  & 0 \\
    \vdots & \ddots & \vdots & \ddots & \vdots\\
    0 & 0 & \frac{\partial \lambda({\xi_{ji}})}{\partial \xi_{ji}} & \dots  & 0 \\
    \vdots & \vdots & \vdots & \ddots & \vdots \\
    0 & 0 & 0 & \dots  & 0
\end{bmatrix}
\end{equation}
\begin{equation}
\frac{\partial \sigma(\xi_{ji})}{\partial \xi_{ij}}=\sigma(\xi_{ji})(1-\sigma(\xi_{ji}))
\end{equation}
\begin{equation}
\frac{\partial \lambda({\xi_{ji}})}{\partial \xi_{ji}} = \frac{1}{2\xi_{ji}}\frac{\partial \sigma(\xi_{ji})}{\partial \xi_{ji}}- \frac{1}{2}\left(\sigma(\xi_{ji})-\frac{1}{2}\right) \frac{1}{\xi_{ji}^2}
\end{equation}

Gradient ascent algorithm cannot be directly applied to constrained optimization problems. There are several procedures that can be applied in constrained problem optimization. The first one involves $\log$ transformation and it was presented in \cite{radosavljevic2010continuous}. The procedure can be further extended by some of the adaptive learning parameter methods. However, in this paper due to large number of parameters, the truncated Newton algorithm for constrained optimization (TNC) was used. More details about TNC can be found in \cite{nocedal2006numerical} and \cite{facchinei2002truncated}. It is necessary to emphasize that the conditional log likelihood is not convex function of parameters $\bm{\alpha}$, $\bm{\beta}$ and $\bm{\xi}$. Because of this, finding a global optimum cannot be guaranteed.

\subsubsection{Learning in GCRFBCnb Model}
Learning in GCRFBCnb model is simpler compared to the GCRFBCb algorithm, because instead of marginalization, the mode of posterior distribution of continuous latent variable $\bm{z}$ is evaluated directly so there is no need for approximation technique. The conditional log likelihood can be expressed as:
\begin{equation}
\underline{{\cal L}}\left(\bm{Y}|\bm{X},\bm{\theta},\bm{\mu}\right)=\log P(\bm{Y}|\bm{X},\bm{\theta},\bm{\mu})=\sum_{j=1}^M \sum_{i=1}^N \log P(y_{ji}|\bm{x},\bm{\theta},\mu_{ji})=\sum_{j=1}^M \sum_{i=1}^N\underline{{\cal L}}_{ji}(y_{ji}|\bm{x},\bm{\theta},\mu_{ji})
\end{equation}
\begin{equation}
\underline{{\cal L}}_{ji}(y_{ji}|\bm{x},\bm{\theta},\mu_{ji})=y_{ji}\log \sigma(\mu_{ji})+(1-y_{ji}) \log \left(1- \sigma(\mu_{ji})\right)
\end{equation}

The derivatives of the conditional log likelihood with respect to $\bm{\alpha}$ and $\bm{\beta}$ are defined as, respectively:
\begin{equation}
\frac{\partial \underline{{\cal L}}_{ji}(y_{ji}|\bm{x},\bm{\theta},\mu_{ji})}{\partial \alpha_k}=\left(y_{ji}-\sigma(\mu_{ji})\right)\frac{\partial \mu_{ji}}{\partial \alpha_{k}}
\end{equation}
\begin{equation}
\frac{\partial \underline{{\cal L}}_{ji}(y_{ji}|\bm{x},\bm{\theta},\mu_{ji})}{\partial \alpha_l}=\left(y_{ji}-\sigma(\mu_{ji})\right)\frac{\partial \mu_{ji}}{\partial \beta_{l}}
\end{equation}
where $\frac{\partial \mu_{ji}}{\partial \alpha_{k}}$ and $\frac{\partial \mu_{ji}}{\partial \beta_{l}}$ are elements of the vectors  $\frac{\partial \bm{\mu_{j}}}{\partial \alpha_{k}}$ and $\frac{\partial \bm{\mu_{j}}}{\partial \beta_{l}}$ and can be obtained by Eqs.~\ref{alpha} and~\ref{beta}, respectively.

In a similar manner TNC or $\log$ transformation gradient ascent algorithms can be used. Additionally, an iterative sequential quadratic programming for constrained nonlinear optimization  can be used, as a result of small number of optimization parameters \cite{boggs1995sequential}.

\section{Experimental Evaluation}
\label{Sec:Experiments}
Both proposed models were tested and compared on synthetic data.
All methods are implemented in Python and experiments were run on Ubuntu server with 128 GB of memory and Intel Xeon 2.9 GHz CPU. All used codes are publicly available.\footnote{https://github.com/andrijaster}

To calculate classification performance of all presented classifiers, the area under ROC curve (AUC) score was used. The AUC score assumes that the classifier outputs a real value for each instance and estimates a probability that for two randomly chosen instances from two different clases the instance from the positive class will have higher value than the instance from the negative class \cite{mohri2018foundations}. A score of 1 indicates perfect classification, whereas score of 0.5 indicates random prediction performance. Aside of AUC score, the lower bound (in case of GCRFBCb) of conditional log likelihood $\underline{{\cal L}}\left(\bm{Y}|\bm{X},\bm{\theta},\bm{\mu}\right)$ and actual value (in case of GCRFBCnb) of conditional log likelihood ${\cal L}\left(\bm{Y}|\bm{X},\bm{\theta}\right)$ of obtained values on synthetic test dataset were also reported.

\subsection{Synthetic Dataset}

The main goal of experiments on synthetic datasets was to examine models under various controlled conditions and show advantages and disadvantages of each. In all experiments on synthetic datasets two different graphs were used (hence $\bm{\beta} \in \mathbb{R}^{2}$) and two unstructured predictors (hence $\bm{\alpha} \in \mathbb{R}^{2}$). In order to generate and label nodes in graph, edge weights $S$ and unstructured predictor values $R$ were randomly generated from uniform distribution. Besides, it was necessary to choose values of parameters $\bm{\alpha}$ and $\bm{\beta}$. Greater values of $\bm{\alpha}$ indicate that model is more confident about performance of unstructured predictors, whereas for the larger value of $\bm{\beta}$ model is putting more emphasis on the dependence structure of output variables.

For generated $S$, $R$, and given parameters $\bm{\alpha}$ and $\bm{\beta}$, probabilities of outputs are obtained and labeling is performed according to the threshold of 0.5. The complete dataset with unstructured predictors, dependence structure  and labeled nodes is used for optimizing parameters $\bm{\alpha}$ and $\bm{\beta}$. Additionally, 20\% of all data was used for testing and 80\% for training procedure.

\subsubsection{Prediction Performance Evaluation}

The main goal of this experiment is to evaluate how the selection of parameters $\bm{\alpha}$ and $\bm{\beta}$ in data generating process affects prediction performance of GCRFBCb and GCRFBCnb. Six different values of parameters $\bm{\alpha}$ and $\bm{\beta}$ were used. The values of parameters were separated in three distinct group:
\begin{enumerate}
\item The first group, in which $\bm{\alpha}$ and $\bm{\beta}$ have similar values. Hence, unstructured predictors and dependence structure between outputs have similar importance.
\item The second group, in which $\bm{\alpha}$ has higher values compared to $\bm{\beta}$, which means that model is putting more emphasis on unstructured predictors in comparison with dependence structure.
\item The third group, in which $\bm{\beta}$ has higher values compared to $\bm{\alpha}$, thus model is putting more emphasis on dependence structure and less on unstructured predictors.
\end{enumerate}

Along with the AUC and conditional log likelihood, norm of the variances of latent variables (diagonal elements in the covariance matrix) is evaluated and presented in Table~\ref{Tab1}. It can be noticed, in cases where norm of the variances of latent variables is insignificant, both models have equal performance considering AUC and conditional log likelihood $\underline{{\cal L}}\left(\bm{Y}|\bm{X},\bm{\theta}\right)$.  This is the case when values of parameters $\bm{\alpha}$ used in data generating process are greater or equal than values of parameters $\bm{\beta}$. Therefore, condtional distribution $P(\bm{y,z}|\bm{x},\bm{\theta})$ is highly concentrated around mean value and MAP estimate is a satisfactory approximation.
However, when data were generated from distribution with significantly higher values of $\bm{\beta}$, compared to $\bm{\alpha}$ the GCRFBCb performs significantly better than GCRFBCnb. For the larger values of variance norm this difference is also large. It can be concluded that GCRFBCb has at least equal prediction performance as GCRFBCnb. Also, it can be argued that the models were generally able to utilize most of the information (from both features and the structure between outputs), which can be seen through AUC values.

\begin{table}[]
\centering
\captionsetup{justification=centering}
\caption{Comparison of GCRFBCb and GCRFBCnb prediction performance for different values of $\bm{\alpha}$ and $\bm{\beta}$, as measured by AUC, log likelihood, and norm of diagonal elements of the covariance matrix}
\label{Tab1}
\begin{tabular}{|c|c|c|c|c|c|c|}
\hline
\textbf{No.} & \textbf{Parameters}                                          & \multicolumn{3}{c|}{\textbf{GCRFBCb}}     & \multicolumn{2}{c|}{\textbf{GCRFBCnb}} \\ \hline
             &                                                              & \textbf{AUC} & \textbf{$\underline{{\cal L}}\left(\bm{Y}|\bm{X},\bm{\theta}\right)$} & \textbf{$\norm{\bm{\sigma}}_2$} & \textbf{AUC}       & \textbf{${\cal L}\left(\bm{Y}|\bm{X},\bm{\theta}\right)$}       \\ \hline
\textbf{1}   & \textbf{\begin{tabular}[c]{@{}l@{}}$\bm{\alpha}=\left[5, 4 \right]$\\ $\bm{\beta}=\left[5, 22 \right]$\end{tabular}} & 0.812       & -71.150    & 0.000       & 0.812             & -71.151          \\ \hline
\textbf{2}   & \textbf{\begin{tabular}[c]{@{}l@{}}$\bm{\alpha}=\left[1, 18 \right]$\\ $\bm{\beta}=\left[1, 18 \right]$\end{tabular}} & 0.903       & -75.033    & 0.001       & 0.902 & -75.033          \\ \hline
\textbf{3}   & \textbf{\begin{tabular}[c]{@{}l@{}}$\bm{\alpha}=\left[22, 21 \right]$\\ $\bm{\beta}=\left[5, 22 \right]$\end{tabular}} & 0.988       & -83.957    & 0.000       & 0.988            & -83.957          \\ \hline
\textbf{4}   & \textbf{\begin{tabular}[c]{@{}l@{}}$\bm{\alpha}=\left[22, 21 \right]$\\ $\bm{\beta}=\left[0.1, 0.67 \right]$\end{tabular}} & 0.866       & -83.724    & 0.000            & 0.886             & -83.466          \\ \hline
\textbf{5}   & \textbf{\begin{tabular}[c]{@{}l@{}}$\bm{\alpha}=\left[0.8, 0.5 \right]$\\  $\bm{\beta}=\left[5, 22 \right]$\end{tabular}} & 0.860       & -83.353    & 34.827      & 0.817              & -84.009          \\ \hline
\textbf{6}   & \textbf{\begin{tabular}[c]{@{}l@{}}$\bm{\alpha}=\left[0.2, 0.4 \right]$\\ $\bm{\beta}=\left[1, 18 \right]$\end{tabular}} & 0.931       & -70.692    & 35.754      & 0.821             & -70.391           \\ \hline
\end{tabular}
\end{table}

\subsubsection{Runtime Evaluation}
The computational and memory complexity of GCRFBCnb during learning and inference is same as time complexity of standard GCRF \cite{radosavljevic2014neural}. If the training lasts T iterations, overall complexity of GCRF is $O(TN^3)$. However, this is the worst case performance and in case of sparse precision matrix, this can be reduced to $O(TN^2)$. The additional memory complexity of GCRF is negligible, which holds for GCRFBCnb, too.

However, in the case of GCRFBCb memory complexity during training is $O(M)$ due to dependency of variational parameters on the number of instances. Computational complexity is also higher -- $O(TMN^3)$, which can also be reduced to $O(TMN^2)$ in case of sparse precision matrix.

\begin{figure}[H]
    \centering
    \captionsetup{justification=centering}
        \includegraphics[angle=0, width=1\textwidth]{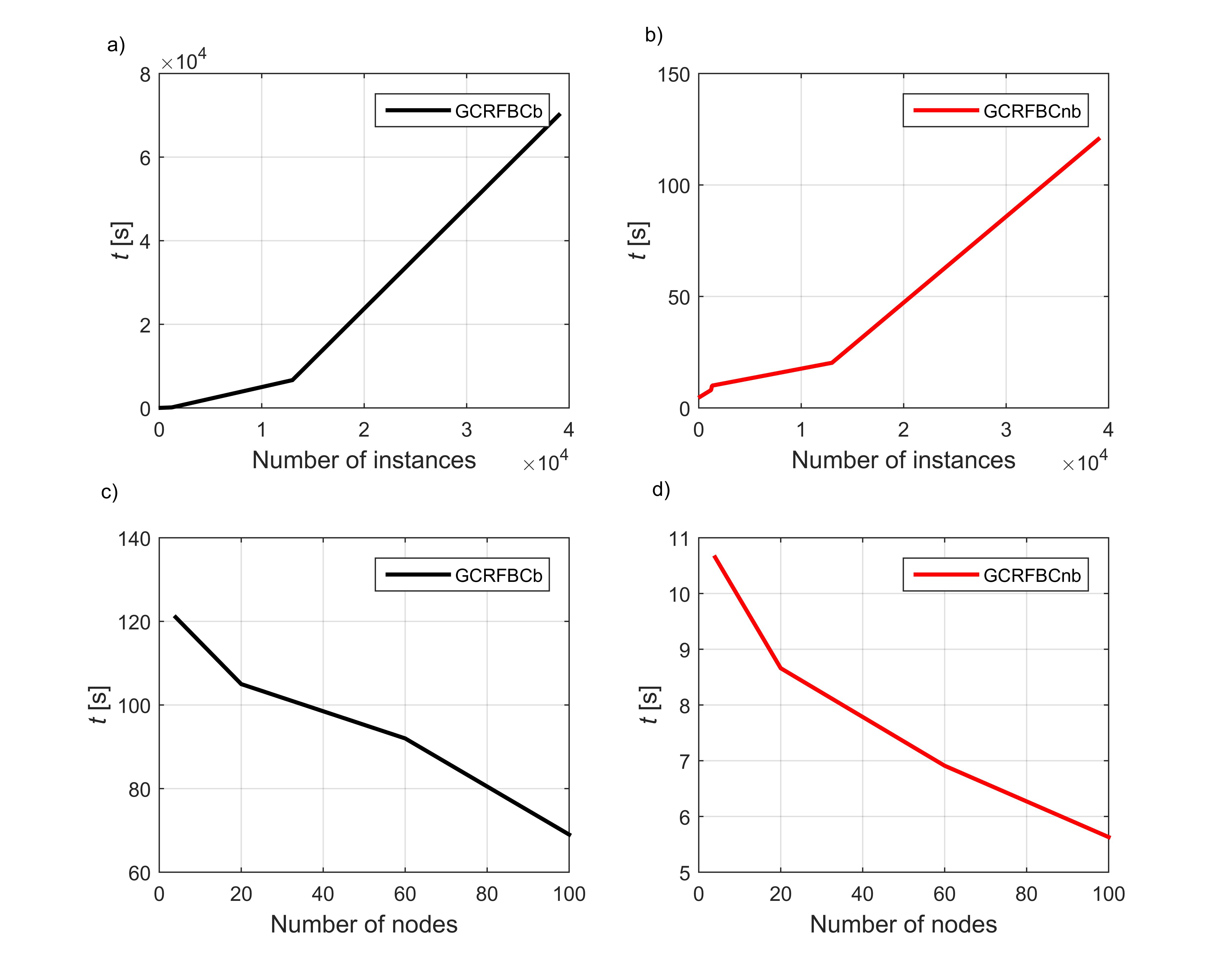}
        \caption{The computational time of GCRFBCb and GCRFBCnb with respect to number of instances and nodes}
        \label{fig:Fig4}
\end{figure}

The following speed tests of GCRFBCb and GCRFBCnb were conducted on synthetically generated data with varying numbers of parameters and nodes. In Figs.~\ref{fig:Fig4}(a) and \ref{fig:Fig4}(b) the computation time of both models with respect to number of instances is presented. The number of nodes in both models is 4. The larger number of instances have significant impact on increase of computation time. 
Figs.~\ref{fig:Fig4}(c) and \ref{fig:Fig4}(d) present computation time with respect to number of nodes, while holding constant value of product of number of instances and nodes (i.e. total number of values of $y$). It can be seen that while holding constant value of products of number of instances and nodes, that computational time increases faster with larger number of instances compared to larger number of nodes.

\section{Conclusion}
\label{Sec:Conclusion}

In this paper, a new model, called Gaussian conditional random fields for binary classification (GCRFBC) is presented. The model is based on latent GCRF structure, which means that intractable structured classification problem can become tractable and efficiently solved. Moreover, improvements previously applied to regression GCRF can be easily extended to GCRFBC. Two different variants of GCRFBC were derived: GCRFBCb and GCRFBCnb. Empirical Bayes (marginalization of latent variables) by local variational methods is used in optimization procedure of GCFRBCb, whereas MAP estimate of latent variables is applied in GCRFBCnb. Based on presented methodology and obtained experimental results on synthetic datasets, several key finding can be summarized:
\begin{itemize}
\item Both models GCRFBCb and GCRFBCnb have better prediction performance compared to the unstructured predictors
\item GCRFBCb has better performance considering AUC score and lower bound of conditional log likelihood $\underline{{\cal L}}\left(\bm{Y}|\bm{X},\bm{\theta}\right)$ compared to GCRFBCnb, in cases where norm of the variances of latent variables is high. However, in cases where norm of the variances is close to zero, both models have equal prediction performance.
\item Due to high memory and computational complexity of GCRFBCb compared to GCRFBCnb, in cases where norm of the variances is close to zero, it is reasonable to use GCRFBCnb. Additionally, the trade off between complexity and accuracy can be made in situation where norm of the variances is high.
\end{itemize}

Further studies should address extending GCRFBC to structured multi-label classification problems, and lower computational complexity of GCRFBCb by considering efficient approximations.

\section*{Acknowledgements}
This research is partially supported by the Ministry of Science, Education and Technological Development of the Republic of Serbia grants OI174021, TR35011 and TR41008. The authors would like to express gratitude to company Saga d.o.o Belgrade, for supporting this research.

\bibliography{literatura}
\bibliographystyle{apalike}
\end{document}